\pdfoutput=1

\documentclass[11pt]{article}

\usepackage[]{EMNLP2023}

\usepackage{times}
\usepackage{latexsym}
\usepackage{graphicx} 

\usepackage[T1]{fontenc}

\usepackage[utf8]{inputenc}

\usepackage{microtype}

\usepackage{inconsolata}

\title{Less is More: A Lightweight and Robust Neural Architecture for Discourse Parsing}

\author{Ming Li \\
  Texas A\&M University \\
  \texttt{liming@tamu.edu} \\\And
  Ruihong Huang \\
  Texas A\&M University \\
  \texttt{huangrh@cse.tamu.edu}}

\begin{document}
\maketitle
\begin{abstract}
Complex feature extractors are widely employed for text representation building. 
However, these complex feature extractors make the NLP systems prone to overfitting especially when the downstream training datasets are relatively small, which is the case for several discourse parsing tasks. 
Thus, we propose an alternative lightweight neural architecture that removes multiple complex feature extractors and only utilizes learnable self-attention modules to indirectly exploit pretrained neural language models, in order to maximally preserve the generalizability of pre-trained language models. 
Experiments on three common discourse parsing tasks show that powered by recent pretrained language models, the lightweight architecture consisting of only two self-attention layers obtains much better generalizability and robustness. 
Meanwhile, it achieves comparable or even better system performance with fewer learnable parameters and less processing time. 
\end{abstract}

\section{Introduction}

Discourse Parsing derives a form of discourse structure that may consist of roles of individual sentences, relations between sentences, or relations between a sentence and a larger text unit, and the derived discourse are widely useful for many NLP applications \cite{yu-etal-2020-better, meyer-popescu-belis-2012-using, ji-etal-2016-latent}. 
We focus on three discourse parsing tasks in this paper, news discourse profiling \cite{choubey-etal-2020-discourse}, RST (Rhetorical Structure Theory) style and PDTB (Penn Discourse Treebank) style discourse parsing tasks \cite{mann1988rhetorical,prasad2008penn}.

\begin{figure}[tb]
\centering 
\includegraphics[width=0.4\textwidth]{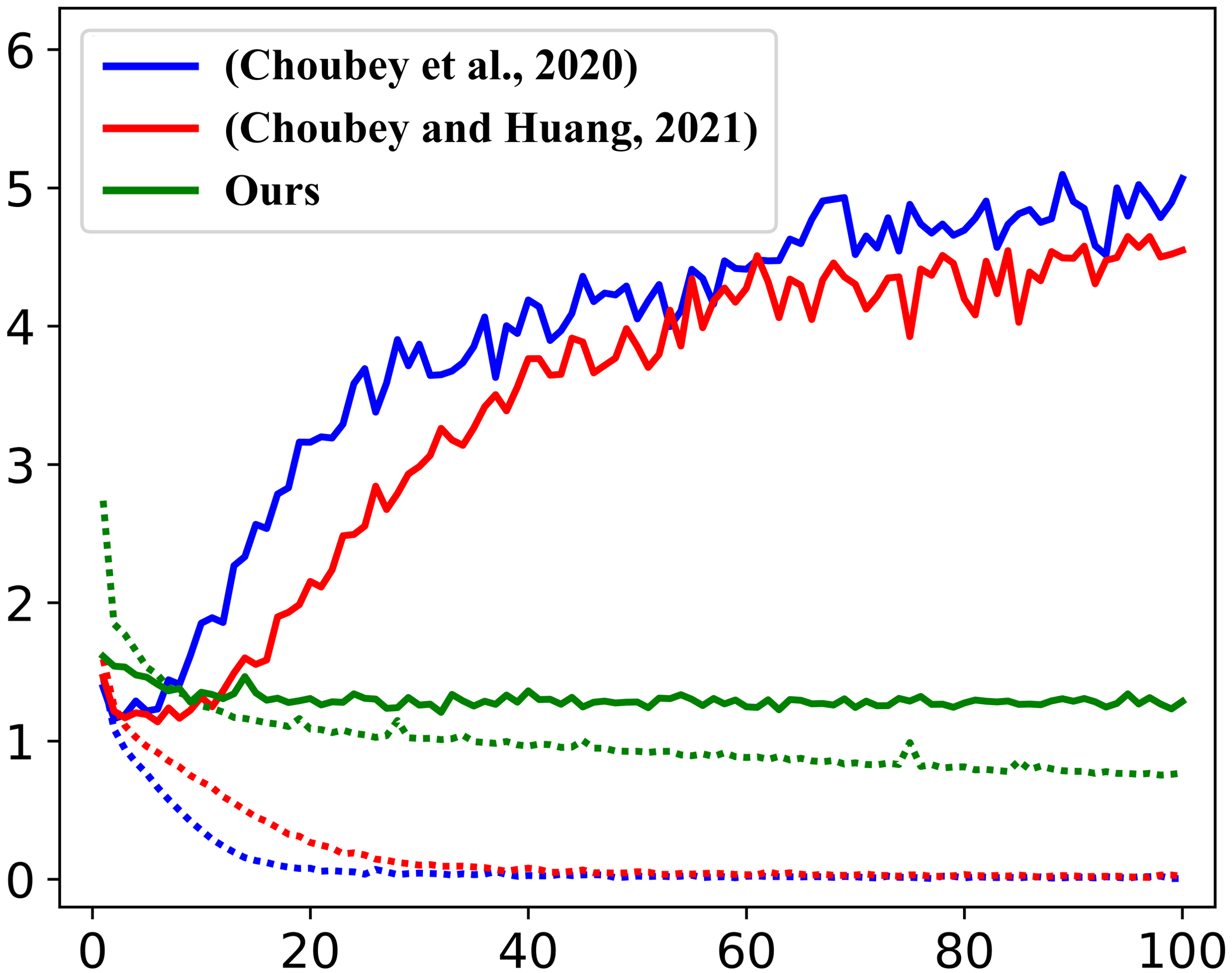} 
\caption{
Training loss curves (dotted lines) and validation loss curves (solid lines) of news discourse profiling models, all under the same training and evaluation settings. 
The horizontal axis represents the number of epochs and the vertical axis represents the averaged loss value. 
For the two previous models, the differences between their training loss and validation loss grow dramatically over training epochs, indicating these models are overfitting the training data. 
In contrast, the validation loss of our simpler model keeps stable and the difference from its training loss is much smaller. 
} 
\label{intro_loss} 
\end{figure}

Like many other NLP systems, neural discourse parsing models often employ complex feature extractors on top of neural language models for text representation building. 
Existing discourse parsing models are mainly based on the heavy use of feature extractors, while the data used for training is limited, which might lead to instability and quick overfitting. 
For instance, in the task of news discourse profiling, \citet{choubey-etal-2020-discourse, choubey-huang-2021-profiling-news} use two Long Short-Term Memory networks (LSTMs) \cite{hochreiter1997long} with self-attention modules to obtain discourse-aware sentence embeddings. 
For RST style discourse parsing, \citet{Kobayashi_Hirao_Kamigaito_Okumura_Nagata_2020, koto-etal-2021-top, yu-etal-2022-rst} use several LSTM or transformer \cite{NIPS2015_1068c6e4} layers for building sentence representations. 
For PDTB style discourse parsing, \citet{dai-huang-2018-improving, bai-zhao-2018-deep, 10.5555/3491440.3491970,10.1145/3485016} also use LSTM layers and transformer layers, some of them further use additional convolution layers or extra memory bank to create sentence embeddings.

However, these complex models are often prone to overfitting, presumably because their complex feature extractors need large datasets to train, but discourse parsing tasks often lack annotated data. 
Take the news discourse profiling task as an example, as shown in Figure \ref{intro_loss}, the two previous models \cite{choubey-etal-2020-discourse,choubey-huang-2021-profiling-news} both overfit the training data and generalize poorly to the unseen validation data. 
This is visually reflected by the large gaps between training loss and validation loss curves. 
Both models can fit perfectly with the training data and achieve near to zero training loss after only several epochs of training while their validation losses are clearly higher and continue to increase dramatically with further training.  
Overfitting with the training data increases the risk that the models perform even worse for future new data and it also makes model tuning sensitive to hyperparameters and randomness.

Thus, we propose an alternative lightweight neural architecture that removes multiple complex feature extractors and only utilizes two well-designed learnable self-attention modules to indirectly exploit pretrained neural language models (PLMs) and build properly integrated sentence embeddings relevant to a particular discourse-level prediction task.  
To maximally keep the generalization ability of PLMs, their weights are fixed all the time and only parameters of self-attention layers are to be learned. 
In these two self-attention layers, the output vectors of each feature extractor are never used directly as inputs to the next layer but only serve as the weights to linearly combine the word embeddings from PLMs. 
By doing so, this simple model is restricted to spoiling the generalization ability of PLMs or fitting with training data. 
Thus it is expected to perform similarly on future new data and mitigates the overfitting and instability problems. 
As shown in Figure \ref{intro_loss}, the validation loss of our simplified model remains stable over the training epochs and the gap between its training and validation losses is kept small, indicating the robustness and generalizability of our model.

We refer to our simple self-attention-based model as LiMNet, where \textit{LiM} stands for \textit{Less is More}. 
In this model, all complex feature extractors are removed from the encoder and two well-designed self-attention modules \cite{bahdanau2014neural, NIPS2015_1068c6e4} are utilized to obtain a local sentence embedding \footnote{We call both Elementary Discourse Unit embeddings in RST and Discourse Unit embeddings in PDTB as sentence embeddings and do not distinguish them strictly.} and sentence-specific complementary information from the discourse. 
These two embeddings of different receptive fields are fused together and then sent to the task-specific predictors. 
It is worth noting that the resulting fused sentence embeddings can be perceived as the linear combination of word embeddings sourced from PLMs, thereby retaining their inherent generalizability.
Extensive experiments and analysis on three discourse parsing tasks including news discourse profiling, RST style and PDTB style discourse parsing, show that LiMNet largely prevents the overfitting problem and gains more robustness. 
Meanwhile, powered by recent PLMs, our simple model achieves comparable or even better system performance with fewer learnable parameters and less processing time. 
This holds significant implications for the ongoing evolution of PLMs, as the fine-tuning of these models becomes increasingly challenging, thereby highlighting the potential of our proposed LiMNet model. 

The contributions of this paper are twofold : 

\begin{itemize}
\item To our best knowledge, we are the first to propose to reduce overfitting by removing complex feature extractors and exploiting pretrained language models in an indirect manner for discourse parsing tasks. 
\item Our model is simple yet effective, it alleviates the potential overfitting problem and becomes more robust to randomness. Besides, it achieves promising performance on three discourse parsing tasks with less processing time, and fewer learnable parameters.
\end{itemize}

\section{Discourse Parsing Tasks and Models}

In this section, we introduce the tasks and models in neural discourse parsing that we focus on and compare the differences between baseline models and our model. 

\subsection{News Discourse Profiling}


News Discourse Profiling \cite{choubey-etal-2020-discourse} is a task focusing on the role of each sentence in news articles \cite{van19865news, van2013news, pan1993framing}. 
It aims to understand the whole discourse at a higher level by identifying the main event with supporting content. 

\begin{figure}[t]
\centering 
\includegraphics[width=0.48\textwidth]{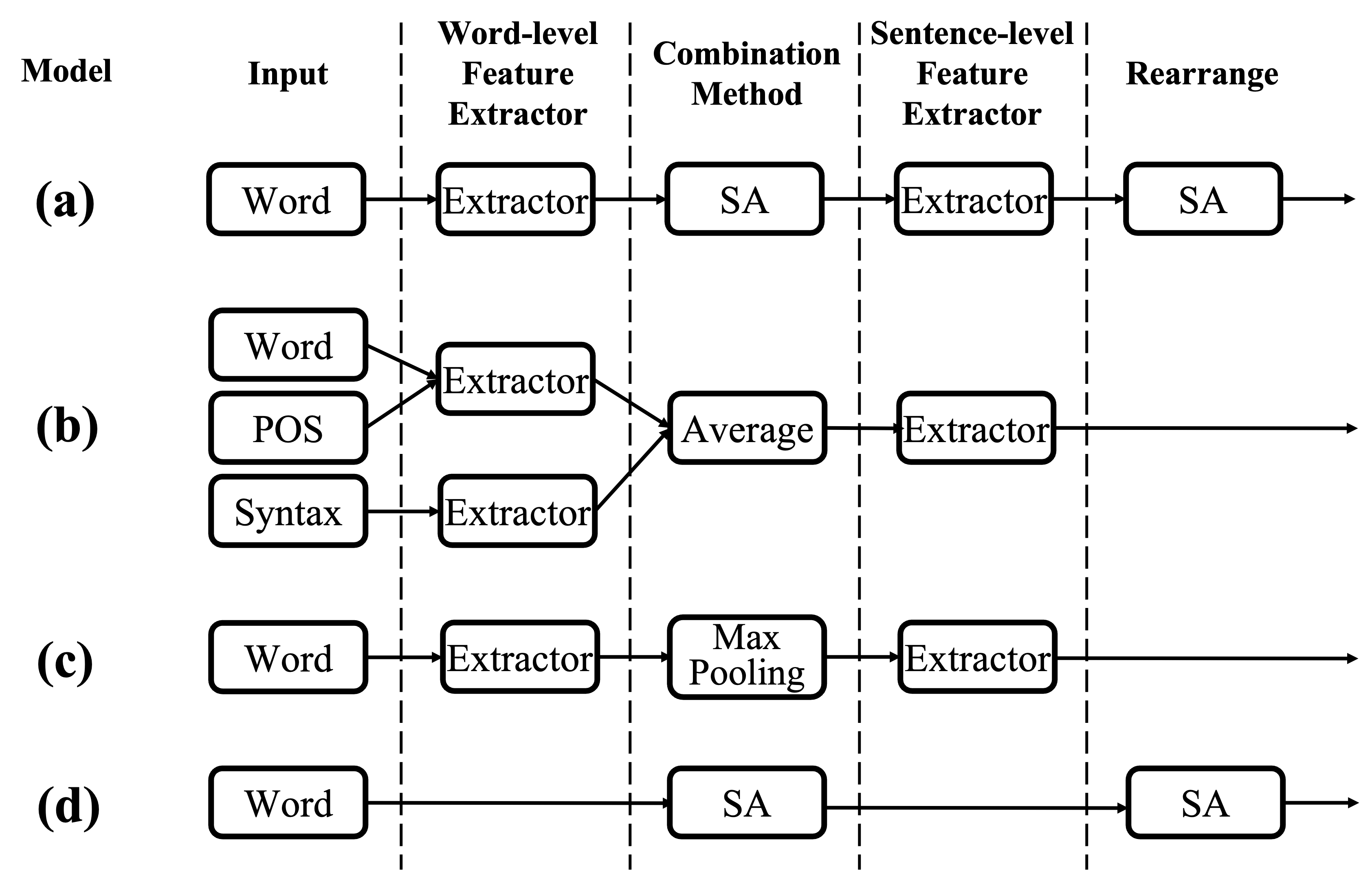} 
\caption{
A brief illustration of feature extractor structures of baseline models and our model. 
(a) is the baseline model for the news discourse profiling task, (b) is the baseline model for the RST discourse parsing task, (c) is the baseline model for the PDTB discourse parsing task, and (d) is our LiMNet. 
\textit{Word} represents input word embeddings, \textit{POS} represents part-of-speech embeddings, and \textit{Syntax} represents syntax features. 
\textit{SA} represents the self-attention module and \textit{Extractor} represents feature extractors utilized by different models. 
All the baseline models (a)(b)(c) utilize word-level and sentence-level feature extractors for sentence representation building, which may lead to severe overfitting. 
Our model (d) removes these additional feature extractors to prevent overfitting. 
} 

\label{all_struct} 
\end{figure}

\vspace{.05in}
\noindent
\textbf{Baseline:}
We use \citet{choubey-etal-2020-discourse} as our baseline model as shown in Figure \ref{all_struct}(a). 
The first feature extractor with a self-attention module is used upon word embeddings to calculate sentence embeddings, and the second is used to obtain document embedding. 
Then the final output vector for each sentence is calculated by the combination of sentence embedding and document embedding. 

\vspace{.05in}
\noindent
\textbf{Updated Baseline:}
For a fair comparison, we utilize word embeddings from T5 \cite{JMLR:v21:20-074} language model as the input while the model structure keeps unchanged. 
Further experiments are conducted in ablation studies to explore the effects of different language models. 

\vspace{.05in}
\noindent
\textbf{Our model:}
As discussed above, introducing extra feature extractors requires more data for training while discourse parsing tasks often lack annotated data, which leads to quick overfitting on training data. 
Thus we remove the word-level and sentence-level Bi-LSTM layers in the baseline and only utilize $2$ self-attention modules to better exploit the pretrained language model, shown in Figure \ref{all_struct}(d). 
Specifically, the first self-attention module is used to build representative local sentence embeddings similar to the baseline. 
The second self-attention module is utilized to calculate the global sentence shifts, aiming to model global contextual information for each sentence. 
This simplification not only alleviates the overfitting problem but also reduces the number of learnable parameters and processing time simultaneously.

\subsection{RST Discourse Parsing}


The purpose of Rhetorical Structure Theory based Discourse Parsing (RST) \cite{mann1988rhetorical,carlson-etal-2001-building} is to present a discourse by using a hierarchical rhetorical tree, where each leaf represents an elementary discourse unit (EDU). 

\vspace{.05in}
\noindent
\textbf{Baseline:}
We use \citet{koto-etal-2021-top} (the LSTM version, their best-performing model structure\footnote{In their paper, the performance of LSTM-based model is better than transformer-based model.}) as our baseline model with a standard decoder as shown in Figure \ref{all_struct}(b). 
The first feature extractor followed by an average pooling is utilized upon the concatenation of word and part-of-speech embeddings in each elementary discourse unit. 
The second feature extractor is utilized upon the syntax features from \citet{DBLP:conf/iclr/DozatM17}. 
The sentence embeddings are obtained by combining the above $2$ embeddings with additional paragraph boundary features. 

\vspace{.05in}
\noindent
\textbf{Updated Baseline:}
We update the baseline to support the T5 language model and discard the POS embeddings, syntax features, and the LSTM applied to syntax features.
In the updated baseline, the first Bi-LSTM layer with average pooling is used upon input T5 word embeddings for local sentence construction. 
And another Bi-LSTM layer is utilized to model contextual relation and the paragraph boundary features are added to the output. 

\vspace{.05in}
\noindent
\textbf{Our model:}
Our model uses the same word embeddings as the updated baseline but replaces all the LSTM layers with self-attention modules, as shown in Figure \ref{all_struct}(d). 
For the discourse unit segmentation step, we use the same decoder as in their experiments.

\begin{figure*}[htb]
\centering 
\includegraphics[width=0.95\textwidth]{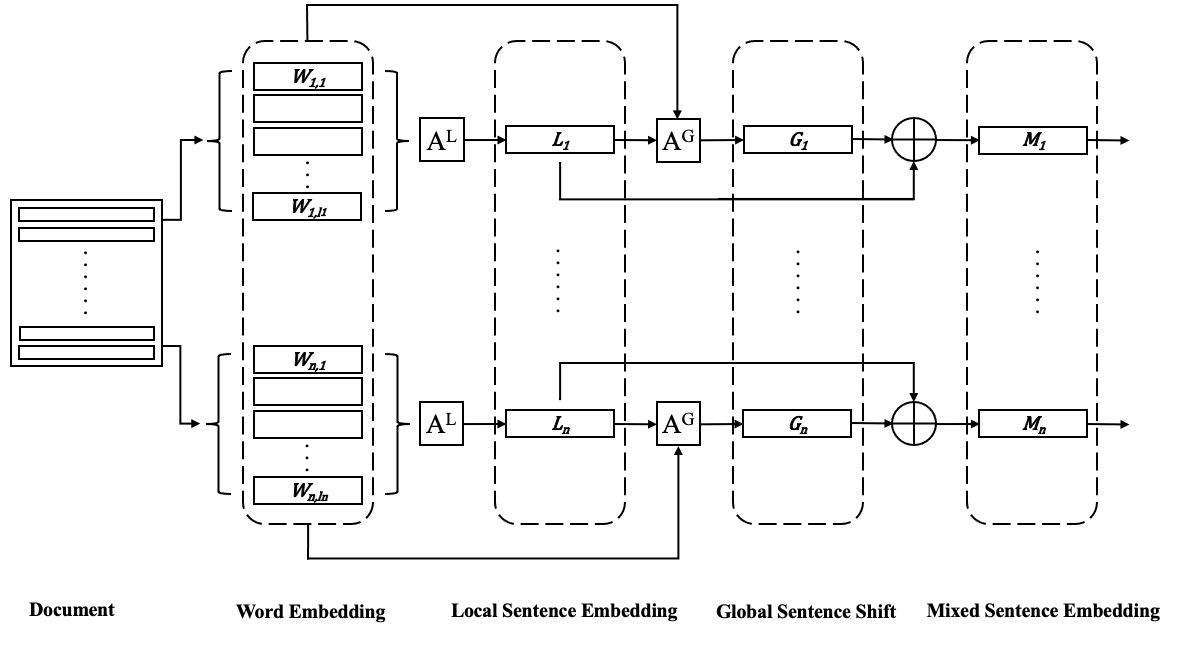} 
\caption{
The structure of our LiMNet. 
$A^L$ represents the self-attention module used for calculating local sentence embeddings and $A^G$ represents the self-attention module used for calculating global sentence shifts. 
The input of $A^G$ contains the $i_{th}$ local sentence embedding and all the word embeddings. 
$\oplus$ represents the addition operation. 
$W_{i,j}$, $L_i$, $G_i$ and $M_i$ represent word embedding, local sentence embedding, global sentence shift and mixed sentence embedding, respectively. 
The mixed sentence embeddings will be later sent to the task-specific predictors. 
} 

\label{model_structure} 
\end{figure*}

\subsection{PDTB Discourse Parsing}


Penn Discourse Treebank based Discourse Parsing (PDTB)\cite{prasad2008penn} focuses more on the relation between $2$ local discourse units. 

\vspace{.05in}
\noindent
\textbf{Baseline:}
Most of the PDTB models only consider the relation between adjacent arguments and neglect the global contextual relation of the whole discourse. 
\citet{dai-huang-2018-improving} is the first context-aware method that models the inter-dependencies between discourse units at the paragraph level. 
The input of this model is the continuous discourse units (DUs) in the same paragraph instead of a single DU pair. 
Thus we choose \citet{dai-huang-2018-improving} as our PDTB baseline where a word-level Bi-LSTM followed by a max pooling is utilized to construct the sentence representations. 
Then a second Bi-LSTM is further utilized upon the obtained sentence embeddings to obtain the final vector output, as shown in Figure \ref{all_struct}(c).

\vspace{.05in}
\noindent
\textbf{Updated Baseline:}
We utilize word embeddings from the T5 language model as the input while the baseline model keeps unchanged. 
Specifically, the whole discourse is sent to the T5 model for capturing global contextual information and the discourse units that need to be predicted are sent to the model at the paragraph level. 

\vspace{.05in}
\noindent
\textbf{Our model:}
Our model uses the same word embeddings as an updated baseline. 
But our model discards all the LSTM layers and utilizes self-attention modules instead, as shown in Figure \ref{all_struct}(d).

\section{LiMNet}
We design LiMNet that removes additional feature extractors and utilizes only self-attention mechanisms to obtain meaningful sentence embeddings indirectly. 
As shown in Figure \ref{model_structure}, the input document will be first sent to the pretrained language model to get all the contextual word embeddings. 
Upon these word embeddings, two self-attention modules are implemented to calculate the local sentence embedding and global sentence shift for each sentence. 
Then the mixed sentence embeddings obtained by adding the previous two embeddings, which can be regarded as the linear combination of pretrained word embeddings, will be sent to the task-specific predictors. 

\vspace{.05in}
\noindent
\textbf{Word Embedding:}
Pretrained language models \cite{peters-etal-2018-deep,devlin-etal-2019-bert} have been widely used in recent NLP tasks, which provide meaningful word embeddings for downstream tasks. 
These language models are mostly trained on large unannotated corpora, which greatly exceed the annotated data used for downstream tasks. 
In neural discourse parsing tasks, we find the finetuning on the language models and using additional complex feature extractors might spoil the generalization ability of pretrained language models and causes performance instability and quick overfitting problems. 
To preserve the generalization ability of pretrained language models, we fix their weights in all the experiments and utilize self-attention modules to better exploit them indirectly. 
In our model, the final sentence embeddings are the linear combination of word embeddings of PLMs, thus preserving their generalization ability. 
As we mainly experiment on discourse-level tasks, we choose T5 \cite{JMLR:v21:20-074} as our language model because it supports long input while many other language models only support inputs up to $512$ tokens.

\vspace{.05in}
\noindent
\textbf{Local Sentence Embedding:}
For discourse-level tasks, getting meaningful sentence embeddings is fundamental. 
Previous models utilize word-level feature extractors followed by a combination method for getting sentence embeddings. 
Striving for building further simplified models, we use only a self-attention module to calculate local sentence embeddings. 
Specifically, two feed forward networks (FFN), with a \textit{tanh} function in between, are utilized upon each word embedding to obtain a scalar representing the importance of this word. 
Then a \textit{softmax} function is applied to get the normalized attention weights and the final local sentence embedding is simply the weighted sum of word embeddings. 

Suppose the $i_{th}$ sentence in discourse has $l_i$ words. 
$W_{i,j}$ represents the word embedding of the $j_{th}$ word of the $i_{th}$ sentence. 
$\alpha^L$ represents the self-attention weight for calculating local sentence embeddings. 
$E^L$ represents the learnable weights for deriving the importance of a word, including $2$ learnable FFN and a \textit{tanh} in between. 
Then the local sentence embedding for the $i_{th}$ sentence can be calculated as follows:

\begin{eqnarray}
    & \alpha^L_{i,j} = softmax(E^L \cdot W_{i,j}) \\
    & L_i = \sum_{j=1}^{l_i} \alpha^L_{i,j}W_{i,j}
\end{eqnarray}

\vspace{.05in}
\noindent
\textbf{Global Sentence Shift:}
Obtaining only local sentence embeddings is insufficient for performing discourse-level tasks. 
Most existing models implement additional discourse-level feature extractors to further incorporate broader discourse-level contexts. 
In addition, each sentence might require specific complementary discourse information to enrich its representation. 
To this end, we implement another self-attention module to learn the global shift for each sentence. 

Specifically, Suppose there are $n$ sentences in a discourse, to obtain the global shift for the $k_{th}$ sentence, differences between its local sentence embedding $L_k$ and all the word embeddings $W_{i,j}$ in the discourse are calculated by subtraction. 
These new vectors represent the shifts between this particular sentence embedding and all other word embeddings. 
Then another self-attention module is implemented upon these sentence shift vectors, and thus the global sentence shift vector $G_k$ of this specific sentence is obtained. 
The global sentence shifts $G_k$ is calculated as follows:
\begin{eqnarray}
    & \alpha^G_{k,i,j} = softmax(E^G \cdot (W_{i,j}-L_k)) \\
    & G_k = \sum_{i=1}^{n} \sum_{j=1}^{l_i} \alpha^G_{k,i,j}W_{i,j}
\end{eqnarray}
where $\alpha^G_{k,i,j}$ represents self-attention weight of word embedding $W_{i,j}$ for calculating global sentence shift of the $k_{th}$ sentence. $E^G$ represents the learnable weights for capturing global sentence shift, including $2$ feed forward networks and a \textit{tanh} function in between. 

The aforementioned equations reveal that both local sentence embeddings and their corresponding global sentence shifts are derived as linear combinations of word embeddings.
Then the final mixed sentence embeddings are obtained by adding global shifts to local sentence embeddings notated as $M_k = L_k + G_k$, which also constitutes a linear combination of word embeddings. 
Such a formulation ensures the maximal preservation of the generalizability inherent in PLMs. These mixed embeddings are then directed toward the task-specific predictors for subsequent processing, thus maintaining the original structure and broad applicability of the PLMs within our model.

\begin{table*}[h]
\centering
\scalebox{0.74}{\begin{tabular}{l|c|c|c|c|c|c|c}
\hline
 & \multicolumn{3}{|c|}{\textbf{Macro}} & \textbf{Micro} & \multicolumn{3}{c}{\textbf{Efficiency}}\\
\hline
\ & Precision & Recall & F1 & F1 & Para(M) & Train(s) & Infer(s) \\ \hline
\ \cite{choubey-etal-2020-discourse} (Baseline)  & $56.9$ & $53.7$ & $54.4 (\pm 0.80)$ & $60.9 (\pm 0.70)$ & $18.02$ & $0.112$ & $0.183$\\ 
\ \cite{choubey-huang-2021-profiling-news} & $58.7$ & $56.4$ & $57.0 (\pm 0.38)$ & $62.2 (\pm 0.59)$ & $30.03$ & $0.139$ & $0.192$\\ \hline

\ \cite{choubey-etal-2020-discourse} w/ T5 (Updated Baseline) & $63.9$ & $61.6$ & $62.3 (\pm 1.25)$ & $66.9 (\pm 0.90)$ & $18.02$ & $0.127$ & $0.205$\\ 
\ \cite{choubey-huang-2021-profiling-news} w/ T5 & $62.5$ & $02.2$ & $61.1 (\pm 0.86)$ & $65.5 (\pm 0.93)$ & $30.03$ & $0.158$ & $0.247$\\ \hline

\ LiMNet (ours) & $\bf 68.2$ & $\bf 63.9$ & $\bf 65.6(\pm 0.42)$ & $\bf 69.7 (\pm 0.25)$ & $3.01$ & $0.110$ & $0.179$ \\ 
 \hline
\end{tabular}}
\caption{\label{tbl:performance}
News discourse profiling results, compared with previous methods in terms of performance and efficiency. 
All the results are averaged over $5$ runs, and the standard deviation for both macro and micro F1 scores are provided in parentheses. 
The original models  \cite{choubey-etal-2020-discourse,choubey-huang-2021-profiling-news} used ELMo language model, thus we updated these models by using T5 pretrained models for direct comparisons, notated with \textit{ w/ T5}. 
\textit{Para}, \textit{Train} and \textit{Infer} represent the number of learnable parameters (in millions), training and inference time (in seconds), respectively. 
The number of learnable parameters does not include parameters of the pretrained language models. 
}
\end{table*}

\section{Evaluation}

\subsection{Dataset}

\vspace{.05in}
\noindent
\textbf{News Discourse Profiling:}
We use the NewsDiscourse dataset \cite{choubey-etal-2020-discourse},  which consists of $802$ news articles ($18,155$ sentences). 
Each sentence in this corpus is labeled with one of eight content types reﬂecting the discourse role it plays in reporting a news story following the news content schemata proposed by Van Dijk \cite{van1985structures, van1988news}. 
For a fair comparison, we use the same division of dataset as \citet{choubey-etal-2020-discourse,choubey-huang-2021-profiling-news}, which has $502$ documents for training, $100$ documents for validation and $200$ documents for testing. 

\vspace{.05in}
\noindent
\textbf{RST Discourse Parsing:}
We use the English RST Discourse Treebank \cite{carlson-etal-2001-building}, which is based on the Wall Street Journal portion of the Penn Treebank \cite{marcus-etal-1993-building}.
It contains $347$ documents for training, and $38$ documents for testing. 
For a fair comparison, we use the same validation set as \citet{koto-etal-2021-top} and \citet{yu-etal-2018-transition}, which contains $35$ documents from the training set. 

\vspace{.05in}
\noindent
\textbf{PDTB Discourse Parsing:}
We use the Penn Discourse Treebank v2.0 \cite{prasad2008penn}, containing $36$k annotated discourse relations in $2,159$ Wall Street Journal (WSJ) articles. 
We use the dataset partition same as \citet{dai-huang-2018-improving} and \citet{rutherford-xue-2015-improving}, where sections $2$-$20$ are training set, sections $21$-$22$ are test set, and sections $0$-$1$ are validation set. 
In this work, we focus on top-level discourse relations. 

\subsection{Implementation Details}

All experiments are implemented in PyTorch platform \cite{paszke2019pytorch} and all the training and inference times are calculated by using NVIDIA GeForce RTX 3090 graphic card. 
We use \emph{t5-large} \cite{JMLR:v21:20-074} from \emph{huggingface} \cite{wolf2019huggingface} as our pretrained language model in all the models. 
The weights of pretrained language model are fixed all the time without finetuning. 
For each model of a different task, we follow the same configuration of the original baselines. 
News Discourse Profiling models are trained using Adam optimizer \cite{kingma2014adam} with the learning rate of $5e-5$ for $100$ epochs. 
RST models are trained using Adam optimizer \cite{kingma2014adam} with the learning rate of $1e-3$, epsilon of $1e-6$, and gradient accumulation of $2$ for $120$ epochs. 
PDTB models are trained using Adam optimizer \cite{kingma2014adam} with the learning rate of $5e-4$ for $100$ epochs. 
The dropout rates \cite{JMLR:v15:srivastava14a} of all the models are set to $0.5$. 
We run each model for $5$ rounds using different seeds and report the averaged performance to alleviate the influence of randomness.

\subsection{Results on News Discourse Profiling}

As shown in Table \ref{tbl:performance}, our model has the best performance on all of the evaluation metrics (macro P/R/F and micro F1 score). 
In addition to the \textit{Baseline} model \cite{choubey-etal-2020-discourse} and its T5 version (\textit{Updated Baseline}), our model also outperforms \citet{choubey-huang-2021-profiling-news} which additionally uses subtopics structures to guide sentence representation building in an actor-critic framework and its T5 version. 
It is impressive that the linear combination of word embeddings is so powerful that it outperforms those models with multiple additional feature extractors. 
Compared with previous models, our model has fewer parameters and less training and inference time, indicating the \textbf{efficiency} of our model. 
The calculation of all sentence embeddings in our model can be done in parallel, leading to large increases in speed. 
In parentheses are the standard deviations of different models on $5$ random runs, where we can see the standard deviation of our model is much less than the previous two models (T5 version), indicating the \textbf{robustness} of our model to randomness. 
In addition, as shown in Figure \ref{rst_pdtb_loss}(a), the \textit{Updated Baseline} is prone to overfitting quickly while our model keeps the training and validation loss similar, indicating the better \textbf{generalization} ability of our model to unseen data used for validation.

\begin{table*}[h]
\centering
\scalebox{0.8}{\begin{tabular}{l|c|c|c|c|c|c|c|c|c|c|c}
\hline
& \multicolumn{4}{|c|}{\textbf{Original Parseval}} & \multicolumn{4}{|c}{\textbf{RST Parseval}} & \multicolumn{2}{|c}{\textbf{Efficiency}}\\ 
\hline
\ & \textbf{S} & \textbf{N} & \textbf{R} & \textbf{F} & \textbf{S} & \textbf{N} & \textbf{R} & \textbf{F} & \textbf{Para(M)} & \textbf{Train(s)} & \textbf{Infer(s)} \\ \hline
\ \cite{hayashi-etal-2016-empirical} & $65.1$ & $54.6$ & $44.7$ & $44.1$
& $ 82.6 $ & $ 66.6 $ & $ 54.6 $ & $ 54.3 $ & - & -& -\\
\ \cite{li-etal-2016-discourse} & $64.5$ & $54.0$ & $38.1$ & $36.6$
& $ 82.2 $ & $ 66.5 $ & $ 51.4 $ & $ 50.6 $ & - & -& -  \\
\ \cite{braud-etal-2017-cross} & $62.7$ & $54.5$ & $45.5$ & $45.1$
& $ 81.3 $ & $ 68.1 $ & $ 56.3 $ & $ 56.0 $ & - & -& -  \\
\ \cite{yu-etal-2018-transition} & $71.4$ & $60.3$ & $49.2$ & $48.1$
& $ 85.6 $ & $ 72.9 $ & $ 59.8 $ & $ 59.3 $  & - & -& -  \\
\ \cite{mabona-etal-2019-neural} & $67.1$ & $57.4$ & $45.5$ & $45.0$ 
& - & - & - & - & - & -& -  \\
\ \cite{Kobayashi_Hirao_Kamigaito_Okumura_Nagata_2020} & - & - & - & - 
& $ 87.0 $ & $ 74.6 $ & $ 60.0 $ & - & - & -& -  \\
\ \cite{zhang-etal-2020-top} & $ 67.2 $ & $ 55.5 $ & $ 45.3 $ & $ 44.3 $ 
& - & - & - & - & - & -& - \\
\ \cite{koto-etal-2021-top} & $ 73.1 $ & \underline{$ 62.3 $} & $ 51.5 $ & $ 50.3 $ 
& \underline{$86.6 $} & $73.7 $ & $ 61.5 $ & $ 60.9 $ & - & -& -  \\
\hline
\ \cite{koto-etal-2021-top}
(Baseline) & $ 72.7 $ & $ 61.7 $ & $ 50.5 $ & $ 49.4$ 
& $86.4 $ & $73.4 $ & $60.8 $ & $ 60.3 $ & $10.16$ & $0.244$ & $3.36$ \\
\ Updated Baseline & $ \bf 75.4 $ & $ \bf 64.1 $ & $ \bf 53.6 $ & $ \bf 52.1$ 
& $ \bf 87.7 $ & $ \bf 75.0 $ & $ \bf 63.1 $ & $ \bf 62.3 $ & $4.79$  & $0.257$ & $2.71$ \\
\ LiMNet (ous) & \underline {$ 73.2 $} & $ 62.0 $ & \underline{$ 51.7 $} & \underline{$ 50.7$ }
& \underline{$86.6 $} & \underline{$73.9 $} & \underline{$61.8 $} & \underline{$ 61.3 $} & $2.29$ & $0.183$ & $2.79$ \\ \hline

\ Deviation (Updated Baseline) & $ 0.09 $ & $ 0.49 $ & $ 0.72 $ & $ 0.64 $ 
& $0.05 $ & $0.27 $ & $ 0.51 $ & $ 0.46 $ & - & -& -  \\
\ Deviation (LiMNet) & $ 0.04 $ & $ 0.07 $ & $ 0.18 $ & $ 0.15 $ 
& $0.01 $ & $0.26 $ & $ 0.14 $ & $ 0.17 $ & - & -& -  \\
\hline
\end{tabular}}
\caption{\label{tbl:rst}
RST discourse parsing results, using original Parseval and RST Parseval metrics. 
S, N, R, F represents Span, Nuclearity, Relation and Full. 
\citet{koto-etal-2021-top} (baseline) represents the LSTM version with the static oracle of the original model, which is used as our RST baseline. 
\textit{Updated Baseline} represents the updated version of its original model to match the use of T5 language model. 
The results of ours and baseline models have averaged over $5$ runs. 
The highest performances are in bold and the second-highest performances are underlined. 
\textit{Para}, \textit{Train} and \textit{Infer} represent the number of learnable parameters, training and inference time, respectively. 
The number of learnable parameters does not include parameters of the language model. 
}
\end{table*} 

\begin{table}[ht]
\centering
\scalebox{0.58}{\begin{tabular}{l|c|c|c|c|c}
\hline
& \multicolumn{2}{|c|}{\textbf{Implicit}} & \multicolumn{3}{|c}{\textbf{Efficiency}} \\ \hline

 & \textbf{Macro} & \textbf{Acc} &  \textbf{Para(M)} & \textbf{Train(ms)} & \textbf{Infer(ms)} \\
\hline
\ \cite{shi-demberg-2019-learning} & $ 46.40 $ & $61.42$
& - & - & - \\ 
\ \cite{Guo_He_Dang_Wang_2020} & $ 47.90 $ & $57.25$
& - & - & - \\ 
\ \cite{dai-huang-2019-regularization} & $ 52.89 $ & $59.66$
& - & - & - \\ 
\ \cite{nguyen-etal-2019-employing} & $ 53.00 $ & - 
& - & - & - \\ 
\ \cite{varia-etal-2019-discourse} & $ 50.20 $ & $ 59.13 $ 
& - & - & - \\ 
\ \cite{wu2020hierarchical} & $ 55.72 $ & $ 65.26 $ 
& - & - & - \\ 
\ \cite{he-etal-2020-transs} & $ 51.24 $ & $ 59.94$
& - & - & - \\ 
\ \cite{kishimoto-etal-2020-adapting} & $ 58.48 $ & $ 65.26 $ 
& - & - & - \\ 
\ \cite{zhang-etal-2021-context} & $ 53.11 $ & - 
& - & - & - \\ 
\hline
\ \cite{dai-huang-2018-improving} & $ 48.69 $ & $ 58.20 $ 
& $1.09$ & $4.5$ & $2.3$ \\
\ Updated baseline & $ 58.48 $ & $ 65.74 $ 
& $1.87$ & $5.1$ & $2.5$ \\
\ LiMNet (ours) & $ \bf 60.69 $  & $\bf 66.60 $ 
& $1.59$ & $3.1$ & $2.1$ \\\hline
\ Deviation (Updated Baseline) & $ 0.81 $  & $0.85 $ 
& - & - & - \\
\ Deviation (LiMNet) & $ 0.32 $  & $0.42 $ 
& - & - & - \\
\hline
\end{tabular}}
\caption{\label{tbl:pdtb}
PDTB discourse parsing results in implicit discourse relation classification. 
\textit{Updated Baseline} represents the updated version of the original baseline model to match the use of T5 language model. 
The results of the models have averaged over $5$ runs. 
\textit{Para}, \textit{Train} and \textit{Infer} represent the number of trainable parameters, training and inference time. 
The number of trainable parameters does not include parameters of the language model. 
The training and inference time does not include the language model time, following the same code structure with our baseline \cite{dai-huang-2018-improving}. 
}
\end{table} 

\subsection{Results on RST Discourse Parsing}

As shown in Table \ref{tbl:rst}, the \textit{Updated Baseline} model achieves the best performance and outperforms the previous methods across all metrics. 
Meanwhile, our model achieves the second-best performance on all the metrics except one and the differences from the best performance are kept low. 
These results are also encouraging considering LiMNet due to its \textbf{efficiency}: it has less than half of the learnable parameters and needs less processing time. 
The last two rows present the standard deviation of \textit{Updated Baseline} and our model on $5$ random runs, where the standard deviations of our model are much less than baseline models, indicating the \textbf{robustness} of our model to randomness. 
In addition, as shown in Figure \ref{rst_pdtb_loss}(b), the \textit{Updated Baseline} is prone to overfitting due to its heavy feature extractors, where its validation loss curve begins to increase dramatically from around the $30_{th}$ epoch while its training loss curve still decreases. 
On the contrary, our model LiMNet keeps the validation loss low throughout the training process, indicating the better \textbf{generalization} ability of our model to unseen data used for validation.

\subsection{Results on PDTB Discourse Parsing}

\begin{figure*}[t]
\centering 
\includegraphics[width=0.85\textwidth]{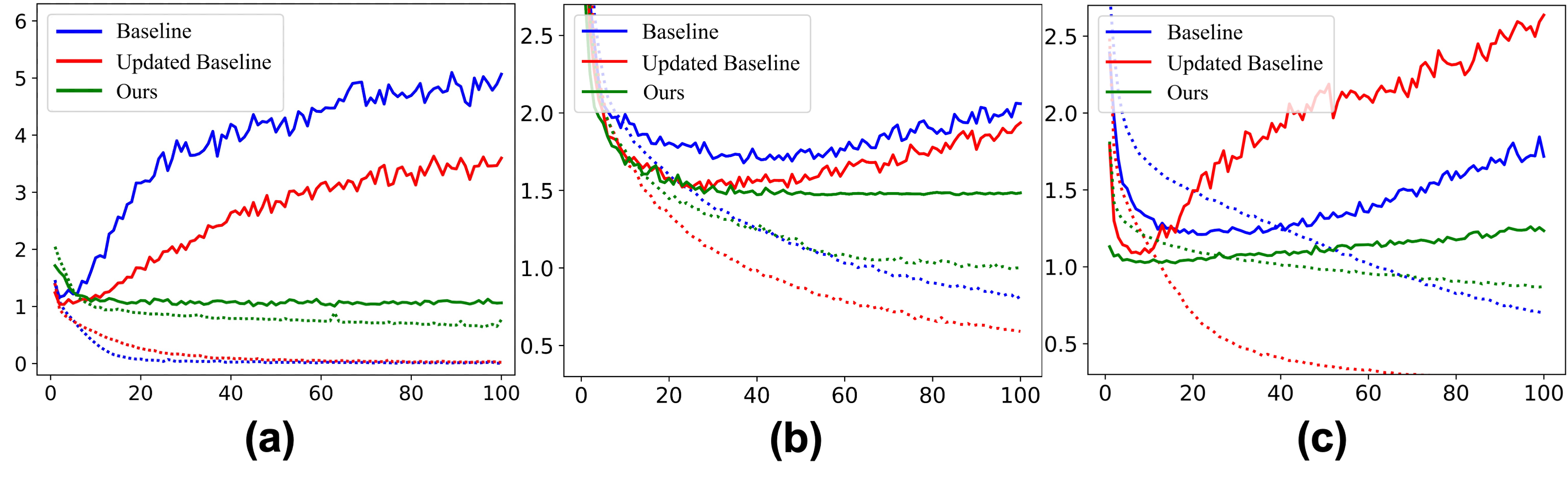} 
\caption{
The training loss curves (dotted line) and validation loss curves (solid line) of (a) news discourse profiling models, (b) RST models and (c) PDTB models. 
The \textit{Baseline} model and \textit{Updated Baseline} model suffer from severe overfitting problems while our model largely reduces this problem. 
} 
\label{rst_pdtb_loss} 
\end{figure*}

As shown in Table \ref{tbl:pdtb}, compared with the baseline model \cite{dai-huang-2018-improving}, the updated baseline greatly improves the performance on implicit discourse relation classification. Meanwhile, LiMNet achieves even better results on both macro F1 and accuracy metrics, in spite of the \textbf{efficiency} of LiMNet since it has fewer parameters than the updated baseline and LiMNet requires less training time and less inference time. 
The last two rows present the standard deviation of \textit{Updated Baseline} and our model on $5$ random runs, where the standard deviations of our model are much less than baseline models, indicating the \textbf{robustness} of our model to randomness. 
As shown in Figure \ref{rst_pdtb_loss}(c), the updated baseline is more prone to overfitting than the baseline model, where its validation loss curve first decreases but then increases even more dramatically. 
However, after removing the feature extractors from the updated baseline, LiMNet largely reduces overfitting, indicating the better \textbf{generalization} ability.

\subsection{Performance Analysis}

From the empirical studies undertaken\footnote{The ablation experiments on model structure, the effect of using different PLM transformer layers on the task of news discourse profiling are further illustrated in the Appendix.}, our simple model architecture outstrips baseline models across multiple aspects on three distinct discourse parsing tasks. The source of these advancements lies in the indirect leveraging of PLMs. 
In contrast to most of the existing studies that resort to the utilization of complex feature extractors, which are intended to transform pretrained embeddings into intricate hidden spaces. Our method focuses on learning how to amalgamate existing word embeddings, which in turn ensures the stability of our model. 
The output representations produced by our model constitute a linear combination of static pretrained word embeddings, thereby safeguarding the inherent generalization capacity of PLMs from degradation. 
Moreover, given the ongoing advancement of PLMs, finetuning these models is becoming increasingly arduous. In such circumstances, a model like ours may offer a viable alternative. As PLMs continue to evolve, our model is poised to persistently preserve and maximally harness their intrinsic capabilities, which should lead to progressively better performance due to improved utilization of PLMs.

\section{Related Work}

\subsection{Overfitting}

In NLP community, transfer learning
\cite{shao2014transfer,weiss2016survey}
and pretraining \cite{erhan2010does,qiu2020pre} are widely used. 
Pretrained language models \cite{pennington-etal-2014-glove,peters-etal-2018-deep,devlin-etal-2019-bert,clark2020electra,Lan2020ALBERT:,Liu2019RoBERTaAR} have been widely used in recent NLP tasks, which provide general and representative word embeddings for downstream tasks. 
Though augmenting parameters increases the performance of various tasks, a great number of parameters may cause overfitting \cite{cawley2007preventing,tzafestas1996overtraining}. 

\subsection{Robustness of Finetuned Models }

The fine-tuning process of these models often displays instability, leading to significantly different performances even with identical settings \cite{DBLP:journals/corr/abs-1909-11299, Zhu2020FreeLB:, Dodge2020FineTuningPL, pruksachatkun-etal-2020-intermediate, mosbach2021on}.
Some research suggests controlling the Lipschitz constant through various noise regularizations \cite{arora2018stronger, Sanyal2020Stable, aghajanyan2021better, hua-etal-2021-noise}. 
Theoretical analysis of fine-tuning paradigms like full fine-tuning \cite{devlin-etal-2019-bert} and head tuning or linear probing \cite{peters-etal-2019-tune, kumar2022finetuning} has been explored, which provide insights into how modifying factors such as the training sample size, iteration number, or learning rate could enhance the stability of the fine-tuning process \cite{wei2021why, kumar2022finetuning, mosbach2021on, hua-etal-2021-noise}.

\section{Conclusions}

We propose to remove additional complex feature extractors and utilize self-attention modules to make good use of the pretrained language models indirectly and retain their generalization abilities. 
Extensive experiments and analysis on three discourse parsing tasks show that our simplified model LiMNet largely prevents the overfitting problem. 
In the meantime, our model achieves comparable or even better system performance with fewer learnable parameters and less processing time. 
Beyond the confines of neural discourse parsing, LiMNet's architecture is task-agnostic. With appropriate task-specific decoders, it can be seamlessly adapted to a variety of tasks, thereby enhancing their generalizability and stability. This underscores the versatility and potential of our neural architecture across a broader spectrum of applications.

\section*{Limitations}

One acknowledged limittaion of our approach lies in its current inability to surpass the state-of-the-art performance benchmarks in RST and PDTB discourse parsing. However, the primary impetus behind this work is not to design a model structure that achieves superior performance. Rather, it is to develop a simple and robust model capable of indirectly harnessing the capabilities of PLMs, with the goal of mitigating potential overfitting issues within the domain of discourse parsing tasks.
Despite not reaching the performance levels of current state-of-the-art systems, our model exhibits superior generalizability and stability, achieved with fewer learnable parameters and less processing time. Looking ahead, as the development of PLMs continues unabated, our model is well-positioned to retain and optimally utilize the intrinsic potential of these models. This approach, we anticipate, will progressively yield improved performance due to the enhanced integration of evolving PLMs.

\bibliography{anthology,custom}
\bibliographystyle{acl_natbib}

\appendix

\section{Ablation Study for Model Components}

A series of ablation experiments were conducted on the news discourse profiling task to examine the functional significance of each module within LiMNet. 

As shown in Table \ref{tbl:model}, \emph{LiMNet w/o Global} represents the model in which the calculation of global sentence shift is removed and the local sentence embeddings are sent to the prediction layer directly. 
This modification led to a minor increment in precision but resulted in a substantial decline in recall, consequently lowering the overall F1 score. Such results underscore the criticality of the global shifts in providing necessary information for the linear combination of word embeddings. 

The variant \emph{LiMNet w/o Local} symbolizes a model wherein the computation of local sentence embeddings is replaced by an average pooling operation. This removal of the sentence-level self-attention module precipitated a significant downturn in both precision and recall metrics. Given that the computation of global sentence shifts is contingent on local sentence embeddings, it is clear that an effective local embedding is paramount to the overall system performance.

\emph{LiMNet w/o Either} represents a model where the computations for both local and global embeddings are excised. In this model, the only learnable component is the FFN in the final prediction layer. The performance of this model presents the inherent representational capabilities of the T5 language model and illuminates the necessity of integrating supplementary neural structures with fixed PLMs. The results demonstrate that in the absence of these additional neural structures, even powerful PLMs such as T5 are unable to yield satisfactory performance when using only their word embeddings. 

In summary, the ablation analysis underscores the utility of each self-attention module within LiMNet and affirms its ability to efficiently harness underlying PLMs to generate task-relevant text representations.

\begin{table}[t]
\centering
\scalebox{0.8}{\begin{tabular}{l|c|c|c|c}
\hline
 & \multicolumn{3}{|c|}{\textbf{Macro}} & \textbf{Micro}\\
\hline
\ & Precision & Recall & F1 & F1 \\ \hline
\ LiMNet & $68.2$ & $ \bf 63.9$ & $\bf 65.6$ & $\bf 69.7 $\\ \hline

\ LiMNet w/o Global & $\bf 69.9$ & $59.9$ & $62.6$ & $69.2 $\\ 
\ LiMNet w/o Local & $62.2$ & $55.9$ & $57.9$ & $62.7$\\ 
\ LiMNet w/o Either & $59.5$ & $46.5$ & $48.7$ & $59.3$\\ 

\hline
\end{tabular}}
\caption{\label{tbl:model}
The performance of different model variations. 
All the models here use the pretrained T5 language model and we show the average results over five runs. 
}
\end{table} 

\begin{table*}[ht]
\centering
\scalebox{0.85}{\begin{tabular}{l|c|c|c|c|c|c|c|c}
\hline
& \multicolumn{4}{|c|}{\textbf{Updated Baseline}} & \multicolumn{4}{|c}{\textbf{LiMNet}} \\ \hline
 & \multicolumn{3}{|c|}{\textbf{Macro}} & \textbf{Micro} & \multicolumn{3}{|c|}{\textbf{Macro}} & \textbf{Micro}\\
\hline
\ & Precision & Recall & F1 & F1 & Precision & Recall & F1 & F1 \\ \hline
\ w/ ELMo \cite{peters-etal-2018-deep} & $56.9 $ & $ 53.7 $ & $ 54.4  $ & $ 60.9  $
& $ 58.1 $ & $ 56.1 $ & $ 56.5 $ & $ 62.5 $\\
\ w/ BERT \cite{devlin-etal-2019-bert}& $ 59.4 $ & $ 58.4 $ & $ 58.6 $ & $ 63.5  $
& $ 62.1 $ & $ 56.4 $ & $ 58.0 $ & $ 64.1 $\\
\ w/ RoBERTa \cite{Liu2019RoBERTaAR}& $ 62.9 $ & $ 57.0 $ & $ 58.8 $ & $ 64.9 $
& $ 65.5 $ & $ 56.3 $ & $ 59.2 $ & $ 67.1$\\ 
\ w/ T5 \cite{JMLR:v21:20-074} & $ 63.9 $ & $ 61.6 $ & $ 62.3$ & $ 66.9$
& $ 68.2 $ & $ 63.9 $ & $ 65.6 $ & $ 69.7 $\\
\hline
\end{tabular}}
\caption{\label{tbl:embedding}
The performance of using different pretrained language models. 
\textit{Updated Baseline} represent \citet{choubey-etal-2020-discourse} model of using different pretrained language models. 
All the results are averaged over $5$ runs, and the standard deviation for both macro and micro F1 scores are provided in brackets. 
The weights of these pretrained language models are fixed without finetuning. 
}
\end{table*} 

\section{Effects of Different Language Models}

\begin{figure}[t]
\centering 
\includegraphics[width=0.5\textwidth]{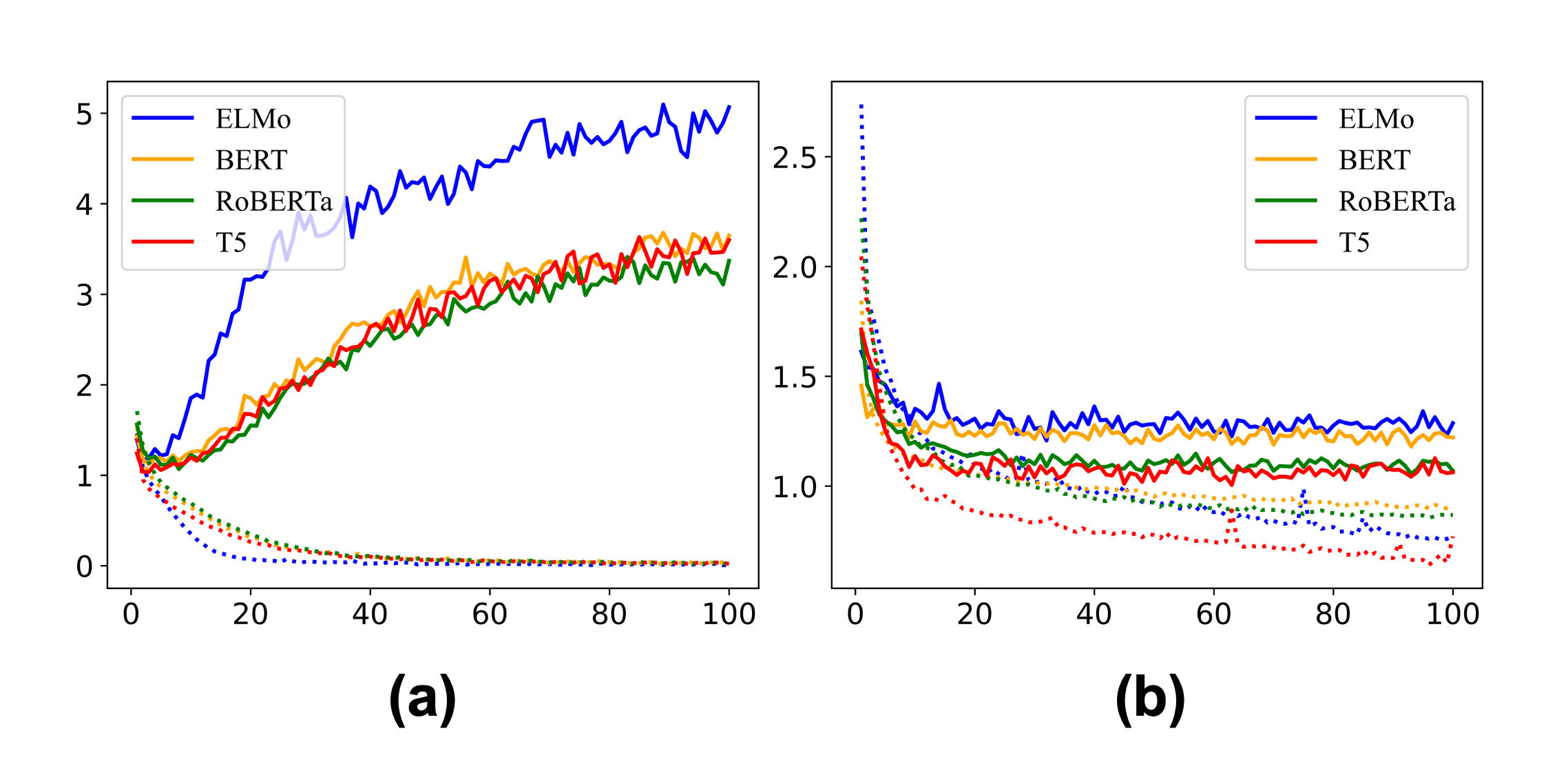} 
\caption{
The training loss curves (dotted line) and validation loss curves (solid line) of (a) baseline models and (b) LiMNet models with different pretrained language models for the News Discourse Profiling task. 
The horizontal axis represents the number of epochs and the vertical axis represents the averaged loss values. 
Across different language models, baseline models are more prone to overfitting while our models are more robust. 
}
\label{ndp_loss} 
\end{figure} 

We also delved deeper into examining the impact of employing various PLMs without finetuning on the task of news discourse profiling. 
Table \ref{tbl:embedding} presents the results of using a series of PLMs for our baseline model and LiMNet. 
Based on the macro F1 metric, our simple neural architecture LiMNet outperforms the baseline model on three pretrained language models, ELMo, RoBERTa and T5, and both models performed comparably on the remaining language model BERT. 
More specifically, LiMNet achieves higher precision across all the language models, but the recall of LiMNet is a little lower on two language models BERT and RoBERTa. Meanwhile, based on the micro F1 metric, LiMNet consistently outperforms the baseline model across all the language models. 
Furthermore, an intriguing observation was that the performance disparity between our LiMNet model and the baseline models widened with the use of superior PLMs. This phenomenon can be reasonably attributed to the fact that our model leverages PLMs indirectly, thus retaining their generalization capabilities while simultaneously being constrained by their inherent capacities.

Furthermore, Figure \ref{ndp_loss} delineates the evolution of the training and validation loss curves for both the baseline models and LiMNet models employing various pretrained language models. Upon inspection, it is observed that for each language model, the disparities between the training and validation loss widen rapidly over the course of the training epochs. This phenomenon is indicative of a propensity for overfitting within the baseline models.
Conversely, the validation loss within the LiMNet models remains relatively stable, and the gaps between their training and validation losses are consistently narrow. These observations affirm the capacity of the LiMNet models to maintain the generalizability inherent in pretrained language models.
The consistency of the generalization capability of our LiMNet model across disparate PLMs underscores its effectiveness in counteracting overfitting issues, thereby further attesting to the robustness of our model architecture.

\section{Effects of Finetuning Language Models}

\begin{figure*}[t]
\centering 
\includegraphics[width=1.0\textwidth]{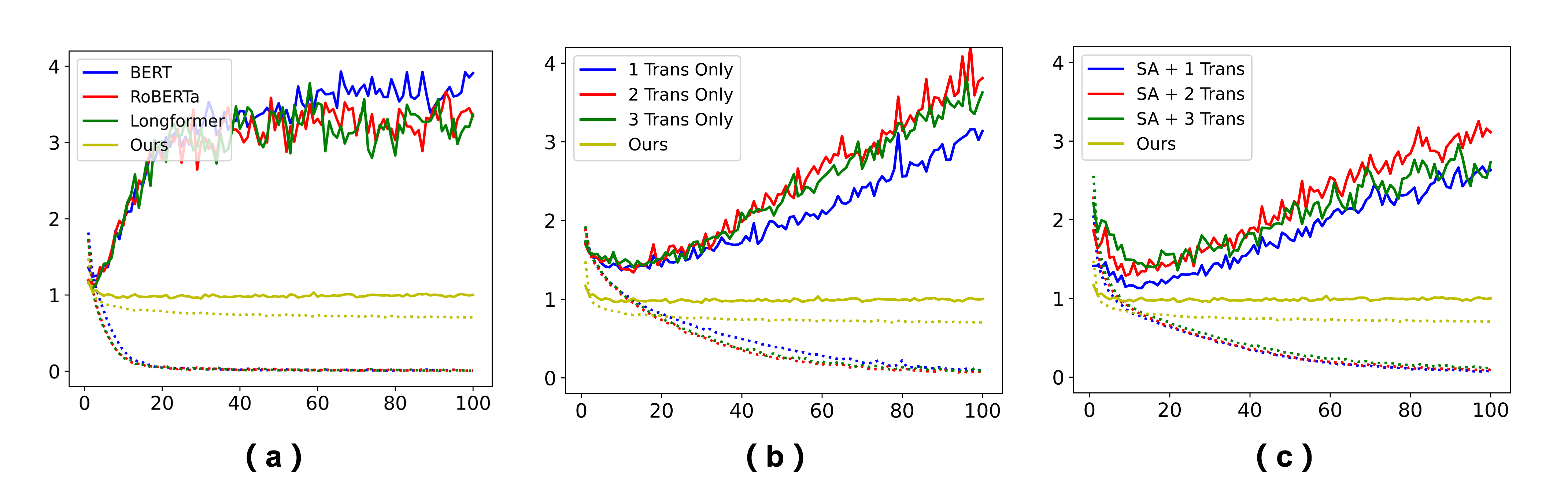} 
\caption{
The training loss curves (dotted line) and validation loss curves (solid line) of (a) models where different language models are unfixed and updated, (b) models where only transformer layers are used upon word embeddings and (c) models where transformer layers are used upon extracted local sentence embeddings. 
} 
\label{appendix} 
\end{figure*}

In this section, we investigate the impact of unfixed pretrained language models that are updated during training for the news discourse profiling task, as demonstrated in Table \ref{tbl:ft}. Compared with experiments involving fixed language models, the finetuning process substantially enhances the final performance metrics.
Although the paradigm of finetuning the entire PLM on downstream tasks is a widely adopted practice, it is susceptible to potentially severe overfitting issues when dealing with tasks characterized by limited data availability. This phenomenon is starkly illustrated in Figure \ref{appendix} (a), which demonstrates the rapid overfitting to training data by models with unfixed PLMs, thereby creating a substantial divergence between training and validation loss.
In comparison to our proposed LiMNet, finetuning PLMs configures their weights to fit specific downstream tasks, potentially compromising their generalization abilities. Furthermore, as PLMs continue to increase in size, finetuning the entirety of these PLMs becomes increasingly challenging. In such a context, our model, which more effectively leverages the fixed PLM, may prove to be increasingly beneficial.\footnote{We do not demonstrate that our approach is superior to the current approaches where the PLMs are updated, especially when comparing the performance scores. However, we would like to highlight our contributions to alleviating the overfitting problem and making the model lightweight by fixing the PLMs. 
}

\begin{table}[t]
\centering
\scalebox{0.7}{\begin{tabular}{l|c|c|c|c}
\hline
 & \multicolumn{3}{|c|}{\textbf{Macro}} & \textbf{Micro}\\
\hline
\ & Precision & Recall & F1 & F1 \\ \hline
\ BERT 
& $58.0$ & $56.8$ & $57.1 (\pm0.82)$ & $63.1(\pm0.25) $\\ 
\ RoBERTa 
& $65.5$ & $64.8$ & $64.9 (\pm0.49)$ & $68.6 (\pm0.82) $\\ 
\ Longformer 
& $65.6$ & $63.1$ & $63.8 (\pm0.47)$ & $68.5 (\pm0.74) $\\  
\hline
\ Ours & $\bf 68.2$ & $\bf 63.9$ & $\bf 65.6(\pm 0.42)$ & $\bf 69.7 (\pm 0.25)$  \\ \hline
\end{tabular}}
\caption{\label{tbl:ft}
The performance of models with different unfixed pretrained language models. 
All the results are averaged over $5$ runs, and the standard deviation for both macro and micro F1 scores are provided in brackets. 
}
\end{table} 

\begin{table}[t]
\centering
\scalebox{0.7}{\begin{tabular}{l|c|c|c|c}
\hline
 & \multicolumn{3}{|c|}{\textbf{Macro}} & \textbf{Micro}\\
\hline
\ & Precision & Recall & F1 & F1 \\ \hline
\ 1 Trans Only & $60.6$ & $58.7$ & $59.1 (\pm0.48)$ & $63.7(\pm0.42) $\\ 
\ 2 Trans Only & $62.6$ & $63.4$ & $62.4 (\pm0.54)$ & $66.1 (\pm0.84) $\\ 
\ 3 Trans Only & $61.8$ & $62.0$ & $61.6 (\pm0.94)$ & $65.5 (\pm0.62) $\\  \hline
\ SA + 1 Trans & $66.2$ & $ 64.2$ & $64.9 (\pm0.52)$ & $ 69.0(\pm0.26) $\\ 
\ SA + 2 Trans & $64.7$ & $64.7$ & $64.3 (\pm0.77)$ & $68.3 (\pm0.39) $\\ 
\ SA + 3 Trans & $64.3$ & $64.7$ & $63.9 (\pm0.63)$ & $67.4 (\pm1.52) $\\  \hline
\ Ours & $\bf 68.2$ & $\bf 63.9$ & $\bf 65.6(\pm 0.42)$ & $\bf 69.7 (\pm 0.25)$  \\ 
\hline
\end{tabular}}
\caption{\label{tbl:trans}
The performance of baseline models with several transformer layers. 
In these settings, the transformer layers are utilized upon local sentence embeddings obtained from the first self-attention module. 
All the results are averaged over $5$ runs, and the standard deviation for both macro and micro F1 scores are provided in brackets. 
All the models here use the pretrained T5 language model. 
}
\end{table} 

\section{Effects of Transformer Layer }

We also provide the experimental results where transformer layers are utilized in replace of our self-attention modules on the task of news discourse profiling as shown in Table \ref{tbl:trans}. 
\textit{x Trans Only} represents the model where $x$ transformer layer with default configuration is utilized upon extracted word embeddings. Then average pooling is implemented to obtain the final sentence embeddings. A considerable performance decline is observed when juxtaposed with our proposed method. Moreover, as shown in Figure \ref{appendix} (b), our model is still more stable than directly using transformer layers. An additional observation is that the incorporation of more transformer layers results in a slight escalation in both performance and standard deviation, thus indicating a reduction in robustness. 
In contrast to a transformer layer, where an FFN is employed to obtain requisite features for the subsequent layer directly, our model's self-attention layer simply learns scalars for composing the word embeddings. This indirect approach circumvents potential adverse effects on word embeddings and thereby preserves their generalizability.

Another group of variant models is denoted as \textit{SA + x Tran}, wherein $x$ transformer layers are applied on top of local sentence embeddings produced by our self-attention module. A significant performance increase is observed when comparing \textit{SA + x Tran} to \textit{x Trans Only}, thereby demonstrating a more effective exploitation of pretrained language models. Nevertheless, our method continues to achieve the highest performance and lowest standard deviation, affirming both the effectiveness of our model in utilizing pretrained language models and the stability of our method. Figure \ref{appendix} (c) provides the loss curves corresponding to each model.

\end{document}